# Proposal for Automatic License and Number Plate Recognition System for Vehicle Identification

Hamed Saghaei

Department of Electrical Engineering, Faculty of Engineering
Shahrekord Branch, Islamic Azad University, Shahrekord - Iran.
Email: h.saghaei@iaushk.ac.ir,
Tel Number: +983833361000-399

*Abstract*— In this paper, we propose an automatic and mechanized license and number plate recognition (LNPR) system which can extract the license plate number of the vehicles passing through a given location using image processing algorithms. No additional devices such as GPS or radio frequency identification (RFID) need to be installed for implementing the proposed system. Using special cameras, the system takes pictures from each passing vehicle and forwards the image to the computer for being processed by the LPR software. Plate recognition software uses different algorithms such as localization, orientation, normalization, segmentation and finally optical character recognition (OCR). The resulting data is applied to compare with the records on a database. Experimental results reveal that the presented system successfully detects and recognizes the vehicle number plate on real images. This system can also be used for security and traffic control.

## I. Introduction

Vehicle's license and number plate recognition (LNPR) system has been an important area of research interest in image monitoring and processing systems [1]. With the advent of high-tech cameras, number plate recognition system has numerous applications for traffic management applications, and especially in the parking lot [2, 3]. LNPR system has many applications such as border crossing control [4], identification of stolen vehicles [5], automated parking attendant [6], red light camera [7], petrol station surveillance, speed enforcement, security. For many of these applications, most of the basic processing algorithms remain the same.

The LNPR system works in three steps, the first step is the detection and capturing a vehicle image, the second one is the detection and extraction of number plate in an image. The third step uses image segmentation technique to get the individual character and optical character recognition (OCR) to recognize the individual character with the help of database stored for each and every alphanumeric character [5, 8, 9].

In LNPR system, the main detection hardware of the first step including some cameras will be installed in places of interest for intersection control, traffic monitoring etc., to identify vehicles that violate traffic laws or to find stolen vehicles. The technique used in some papers is based on pattern matching [10], which is fast and accurate enough for real-time applications and is developed for recognition of license plates with prior knowledge of letters and numbers orientation [11]. Since the orientation and font used for number plates differ in different countries/states/provinces, this algorithm is needed to be modified accordingly keeping its structure intact, if we want to apply this system for recognizing the number plates of those places.

The purpose of this paper is to develop and implement a smart system for optimum use of information and communication technology (ICT) for managing executive organizations as well as forming a database for facilitating decision-making and adopting better staff and strategic planning methods. License plate number is one of the most appropriate information items for identifying a vehicle and its owner in Iran. Vehicle traffic can be controlled through different techniques. If such traffic involves a variety of vehicle models, then installing special technical equipment for each model would not be economical and cannot provide the expected level of security. License plate identification is a useful and the only practical method used in individual countries to ensure security, prevent theft, and manage vehicle fleet.

The rest of the paper is organized as follows: section II will present the system model of the developed LNPR system. Section III will present the LNPR software with detailed information on used algorithms. Section IV presents LNPR system advantages briefly and finally section V will end the paper with the conclusion.

## II. System Model

LNPR system is proposed for monitoring and managing traffic in the parking lots of private and public organizations via identifying vehicle license plate numbers at the parking gate. This system can also be used to identify stolen vehicles on roads. No additional equipment need to be installed on vehicles for operating this system. Fig. 1 shows that the only requirement of this system is installing special cameras for identifying license numbers at the entrance and exit gates of the parking lots. The images taken by these cameras are subsequently processed in a computer. All vehicle traffic information (including the driver's image) is stored in the system database for a long time. Thus, detailed traffic information can be retrieved from different parking gates at different times. Moreover, this system can apply intelligent control in the parking lots through automatic opening of the gate only to authorized vehicles upon recognizing their license numbers according to Fig. 1. Other advantages of the proposed system include online access to information such as the total number of vehicles currently present in the parking lot, the number of authorized vehicles, etc. Moreover, too much

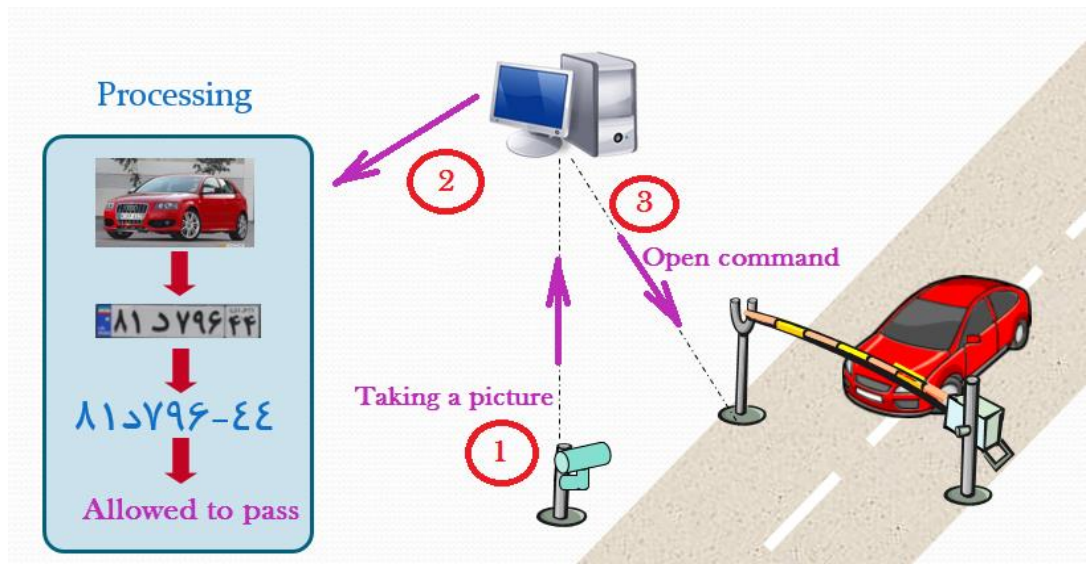

Fig. 1. LNPR system for opening our desired mechanized gate in a parking lot

information of vehicles traffic statistics can be extracted from the system.

The cameras used in the system can be deployed under all weather conditions and are equipped with powerful infrared radiation units for identifying vehicle license plates in absolute darkness. Moreover, these cameras are equipped with thermometers which detect ambient temperature and activate the cooling system of the camera at very high temperatures so as to provide high-quality imaging in the -50 to +70 deg C temperature range. The camera case is built in accordance with the IP66 standard and is fully water and dust proof. The lenses used in the camera provide high-quality image processing capabilities. The proposed comprehensive vehicle traffic control and monitoring system is used for monitoring vehicle access to a specified area. The system normally comprises a camera for monitoring the vehicle path, an identification system for recognizing license plate number used for further identification and control purposes, gates and also traffic lights presented as optional accessories.

The software design and programming with high precision provide great processing speeds and fully reliable system security. The software architecture is so designed that the applications selected by the users can be easily developed with minimum cost in the shortest possible time. The advanced artificial intelligence and neural networks techniques implemented in the processing engine offer users powerful and precise processing capabilities.

In the proposed software, the administrator defines both "white list" for authorized vehicles which can access the site and a "black list" for unauthorized vehicles which are denied entry to that site. The system consequently would allow only the authorized (predefined) vehicles to enter the parking lot, and, meanwhile, prepares reports of the parking lot in terms of the number of current vehicles, time of entry and departure, number of vehicles exiting the area, duration of each vehicle's stay in the area, vehicles which have been denied entry, unidentified vehicles, etc. Implementing the proposed system can provide a higher level of security and safety in the controlled area for the vehicles. The system can be connected to some gates to provide automatic and intelligent opening/closing operation for authorized vehicles. If connected to alarm lights, traffic lights, or smart boards, the system can display predefined messages for specific vehicles (like "welcome", etc.). The user interface of the system is designed for speedy access to system events and can facilitate usage by issuing audio alarms.

### III. Automatic NPR Software

In this section, we introduce LNPR software. a picture was taken by a camera shown in Fig. 2(a) is sent to LNPR software for image processing and number recognition.

To maximize the flexibility of the LNPR system, a modular structure is chosen. In every module image processing algorithm(s) will be implemented. The software uses nine algorithms means nine modules for identifying a license and number plate follows:

1- Plate localization: this algorithm is responsible for finding and isolating the plate on the picture. It is shown in Fig. 2(b)

2- Plate orientation and sizing: this algorithm compensates for the skew of the plate and adjusts the dimensions to the required size that is illustrated in Fig. 2(c).

3- Conversion: Using some conversion image processing techniques, the image can be converted as desired for instance to have a simpler processing of the image, we convert the image from red-green-blue (RGB) layers to gray scale layer demonstrated in Fig. 2(d)

4- Normalization: this algorithm adjusts the contrast and brightness of the image that is illustrated in Fig. 2(e).

5- Edge detection: it is applied to increase the picture difference between the letters and the plate backing. A median filter may also be used to reduce the visual noise on the image. As can be observed in Fig. 2(f), a useless image part similar to the left side of Fig. 2(e) has been omitted.

6- Character segmentation: this algorithm finds the individual characters on the plate and segments them for extra enhancement and also additional lines are deleted as shown in Fig. 2(g).

7- Optical character recognition: it is the electronic conversion of images or printed text into machine-encoded text demonstrated in Fig. 2(h).

8- Syntactical and geometrical analysis: it checks characters and positions against country-specific rules.

9- The averaging of the recognized value over multiple fields/images to produce a more reliable or confident result. Especially since any single image may contain a reflected light flare, be partially obscured or another temporary effect.

*A. Technical Specification of the NPR Processor*

NPR Software has some features including input image, output data and operating systems supported by system core as follows:

Input Image Features:

- BMP or JPEG image formats
- PAL or NTSC video formats
- Average processing time: 60 ms (2 GHz CPU)

Output Data:

- License plate number
- License plate position
- Characters positions
- Driver's image
- License plate color
- Character recognition reliability
- Possibility of identifying all kinds of license plates regardless of the type and place of registration
- Possibility of recognizing license plates for stationary and moving vehicles (up to 180 km/h speed)
- Very high processing speed and precision

Operating Systems Supported by the System Core:

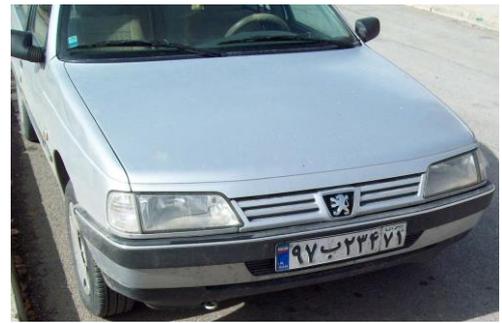
(a) A taken picture by a camera installed at the gate place

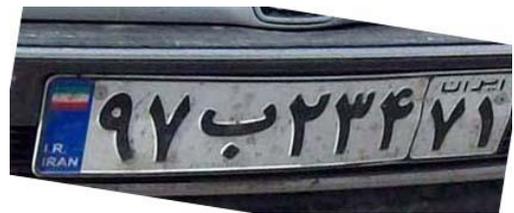
(b) Plate localization

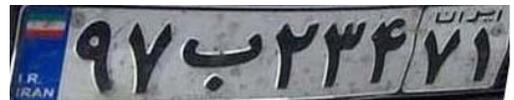
(c) Plate orientation and sizing

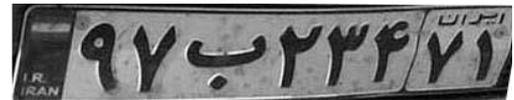
(d) RGB to gray conversion of the Plate

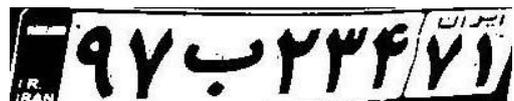
(e) Plate normalization

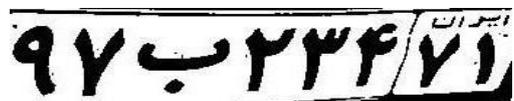
(f) Plate edge detection

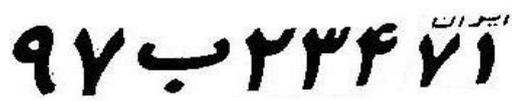
(g) Plate character segmentation

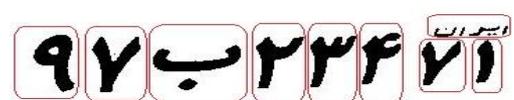
(h) Optical character recognition

Fig. 2. LNPR software with nine image processing algorithms

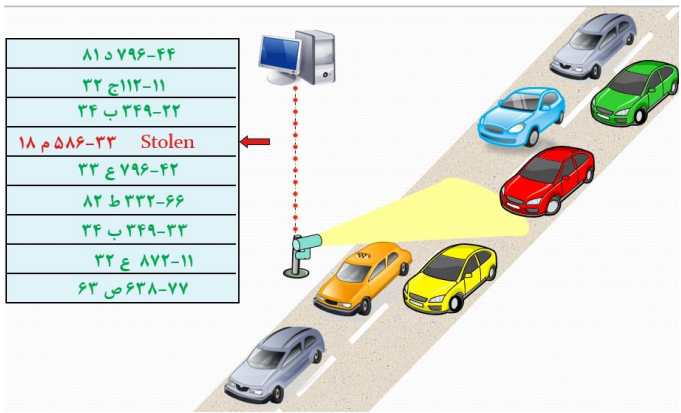

Fig. 3. LNPR system for stolen vehicles identification in highways

- Windows: XP (SP2 and SP3), Vista (SP1), and Server 2003, 2008.
- Linux: Red Hat Enterprise (3-5), Debian (4,5), Fedora Core (7-10), and Open SUSE (7-11).

The main core of the system can be executed under Windows and Linux operating systems. For the high-security and high-sensitivity projects where Windows cannot provide the required security, the Linux-based LPR version can be used for more improved reliability. The received data can be sent using wireless communication system to a defined server. I studied power control process for 3rd and 4th generations of wireless cellular communication systems [12-15].

## IV. LNPR System Advantages

LNPR system has some advantages as follows:

- Easier vehicle's arrivals/departures to/from a parking lot
- Issuing alarms when authorized or unauthorized vehicles try to enter a site
- Preparing updated and instantaneous reports from the situation within the parking lot as well as individual vacant and occupied parking spaces
- Recording and retaining the driver's picture during arrivals/departures
- Archiving vehicle traffic into and out of the area for a very long period of time
- Increasing security in the area
- Presenting various statistical reports about vehicle traffic in and out of the area
- No need to install additional equipment on vehicles for recognizing them (like different cards, etc.)
- Facilitating traffic inflow/outflow during rush hours
- Possibility of displaying or broadcasting messages for certain predefined vehicles
- The possibility of exerting smart control on the gates and traffic lights.

A. *System Capabilities*

- Defining and issuing admission tickets for authorized vehicles to enter the area
- Defining unauthorized vehicles for preventing their entry to the area and also defining a variety of permissions for entering the area
- Defining the authorized day and time of arrival for different permits
- Automatic opening the gate to authorized vehicles immediately upon recognizing their license plate and manual opening of the gate by the guard and issuing permission for specific vehicles as well as registering their license plates
- Defining guards' working shifts and working hours
- Issuing temporary permissions for specific vehicles
- Possibility of taking the driver's picture (optional)
- Possibility of defining specific days and hours for exerting better control and taking images of the driver and the vehicle
- Recording the entry/exit date and time for the vehicles
- Reporting the authorized and unauthorized vehicles parked within the area
- Providing multiple reports regarding the current and past status of vehicle traffic to and from the area
- Web-based monitoring the authorized and unauthorized entries (the number of entries can be increased)
- Storing vehicle traffic information for an unlimited time.

B. *Reports*

The proposed system presents a number of predefined default reports. Based on customer demand, new reports can be added to the system. The default reports include license plate number, vehicle model, driver information, arrivals/departures time, entrance and departure gate numbers, parking duration, and gate shift guard reports.

C. *Applications*

The proposed system can be widely used in the following applications:

- Stolen vehicles identification in roads and highways shown in Fig. 3
- Office areas

- Parking lots of different places such as commercial complex, public parking and etc.
- Passenger terminals
- Airports
- Railway stations
- Shopping centers, auto repair shops, and carwash areas
- Other places where traffic control is required.

## V. CONCLUSION

In summary, this paper presented the automatic vehicle identification system using vehicle license and number plate recognition. The LNPR software of the system uses series of image processing algorithms for number plate recognition and finally identifying the vehicle from the database stored on the PC. This software has been written in .Net C# based on the studied and simulated algorithms in Matlab. The SQL database has been used to store different achieved records of vehicles. We have evaluated the system performance on real images. Both the simulation and practical results revealed that the LNPR system can robustly detect and recognize the vehicle using license plate in different lightening and weather conditions and can be implemented on the entrance of highly restricted areas. The prototype system can be integrated to the intersection surveillance video system for traffic surveying, stolen vehicles, or for some application specific purposes discussed in the paper.